# (SAFE) SMART HANDS: HAND ACTIVITY ANALYSIS AND DISTRACTION ALERTS USING A MULTI-CAMERA FRAMEWORK

**Ross Greer**
**Lulua Rakla**
**Anish Gopalan**
**Mohan Trivedi**, *Faculty Mentor*
University of California San Diego
USA

Student Safety Technology Design Competition: Distraction Prevention and Mitigation

## ABSTRACT

Manual (hand-related) activity is a significant source of crash risk while driving. Accordingly, analysis of hand position and hand activity occupation is a useful component to understanding a driver's readiness to take control of a vehicle. Visual sensing through cameras provides a passive means of observing the hands, but its effectiveness varies depending on camera location. We introduce an algorithmic framework, SMART Hands, for accurate hand classification with an ensemble of camera views using machine learning. We illustrate the effectiveness of this framework in a 4-camera setup, reaching 98% classification accuracy on a variety of locations and held objects for both of the driver's hands. We conclude that this multi-camera framework can be extended to additional tasks such as gaze and pose analysis, with further applications in driver and passenger safety.

## INTRODUCTION

A 2022 CDC study estimated 3,000 deaths due to distracted driving every year, and a significant number of these accidents are due to manual distractions - that is, distractions that involve the driver's hands [1]. Driver distraction contributes to around 65% of safety-critical events (crashes and near crashes) [2]; for example, a study by the Barnes Firm found that eating or drinking while driving increases the chances of getting in a car accident by 80%, and a study by Edgar Snyder & Associates found 25% of car accidents are caused by texting and driving.

Furthermore, given recent consumer adoption of early-stage autonomy in vehicles, driver hand activity has been shown to lead to various incidents even in these semi-autonomous vehicles. Drivers show high propensity to engage in distracting activities when supported by automation [3], and drivers have slower control takeover times when engaged in secondary tasks [4]. Moreover, it is important to consider the controllability of transitions when the driver must take manual control of semi-autonomous vehicles, as drivers demonstrate an inability to handle these control transitions safely when occupied with non driving-related tasks, often involving the hands [5]. Thus, it is evident that monitoring hand activity is critical to ensuring driver safety due to the frequency of accidents that involve the driver's hands.

For both privacy and utility, most current commercial ADAS utilize only outside-facing cameras, and those which do utilize inside-facing cameras tend to support gaze-related alerts, ignoring the criticality of hands. For instance, Honda vehicles alert the driver if they are veering off-lane or driving incorrectly through outward-facing cameras. Current Subaru Outback and Cadillac Super Cruise vehicles offer inside-facing infrared cameras to monitor driver attentiveness and alert if eyes are off the road. A single camera can capture driver activity, but with performance tradeoffs between tasks. For example, a camera aligned to view the eyes will not have an optimal view of the hands, and vice versa.

Moreover, commercial systems which monitor the driver's hands do so without cameras; for example, Tesla's autopilot requires hands to apply pressure to the wheel, and the Hyundai Santa Fe uses steering wheel sensors to monitor the driver's hands. While this provides a framework for determining whether the driver is alert or distracted based on their hand activity, it provides insufficient information to determine what the driver is doing. These torque sensors are effective for determining if the hands are on or off the wheel, but they cannot distinguish between different hand activities taking place off of the wheel, activities which are critical for estimating important metrics like driver readiness and take-over time. Hand locations and held objects imply hand activities, crucial to inferring a driver's state, and this information is lost when reduced to hands-on-wheel and hands-off-wheel.

Furthermore, to reduce cost and minimize driver obtrusiveness, many commercial vehicles contain only one or two interior-facing cameras to monitor driver activity, generally located either above the rearview mirror or above the dashboard display. However, different situations may call for different camera placements for various tasks, so while one view may be ideal for a particular task within design constraints, this view may sacrifice a complete view of a different driver aspect and may not offer redundancies if a camera is obstructed or blocked. For example, an ideal hand view (taken from above the driver) would not be suitable for assessing a driver's eyegaze, but a camera that can see the driver's eyes may also have at least a partial view of the driver's hands.

We have designed SMART Hands (Safe, Multiview Activity Recognition by Tracking Hands), a framework for multiview analysis of driver hand activity which can be applied to arbitrary sets of multi-camera perspectives, addressing the issue of effectively sharing information for a complete and accurate analysis of driver hand activity.

**RELATED RESEARCH**

A driver with their hands off the wheel may be manually, visually, or cognitively distracted [5], and Rangesh et al. [4] and Deo & Trivedi [6] show that driver hand activity is the most important component of models for prediction of driver readiness and takeover time, two metrics critical to safe control transitions in autonomous vehicles. Such driver-monitoring models take hand activity classes and held-object classes as input, among other components, as illustrated in Figure 1. These classes can be inferred from models such as HandyNet [7] and Part Affinity Fields [8], using individual frames of a single camera view as input. Critically, this view is taken to be above the driver, centered in the cabin and directed towards the lap– a typically unobstructed view of the hands.

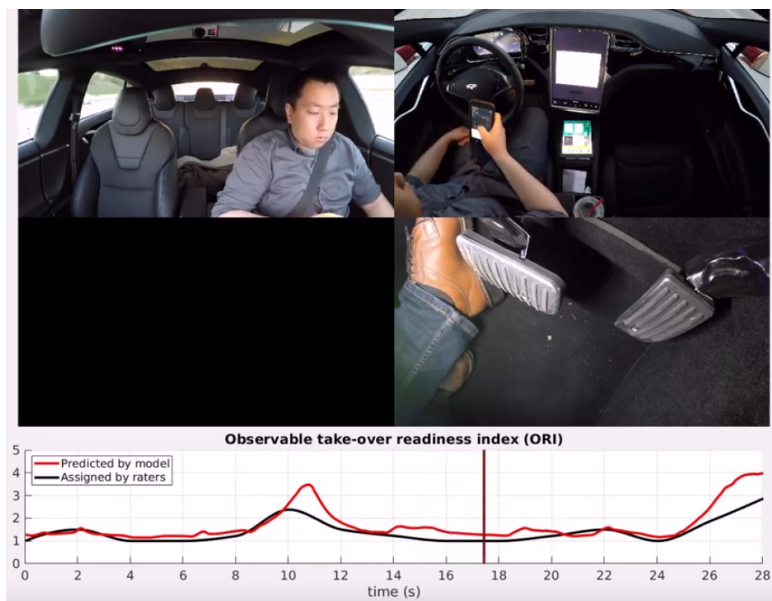

*Figure 1. The Observable Readiness Index uses hand activity features to assess a driver's takeover readiness.*

Initial systems for classification of driver activity and identification of driver distraction used traditional machine learning approaches; for instance, Ohn-Bar et al. demonstrated systems that utilize both static and dynamic hand activity cues in order to classify activity in one of three regions [9] and extracts various hand cues in ROI and fuses them using an SVM classifier [10]. More recent works expand on the aforementioned classifiers and utilize deep learning in order to identify and classify driver distraction in a more robust manner. Eraqi et al., among others, have developed systems that operate in real time to identify driver distraction in a CNN-based localization method [11]. Shahverdy et al. also use a CNN-based system in order to differentiate between driving styles (normal, aggressive, etc.) in order to alert the driver accordingly [12]. Building on this, Weyers et al. demonstrate a system for driver activity recognition based on analysis of key body points of the driver and a recurrent neural network [13], and Yang et al. further build on this work and demonstrate a spatial and temporal-stream based CNN to classify a driver's activity and the object/device causing driver distraction [14]. A comprehensive survey outlining the current driver behavior analysis using in-vehicle cameras was done by Wang et al. [15]

The SMART Hands framework brings two benefits: increased flexibility in field-of-view for the individual component cameras, and increased accuracy in classification. Both benefits arise from the ability of SMART Hands to reason between views, allowing occluded or otherwise compromised images from one view to be substantiated by images from additional views in cooperation.

**METHODS**

The SMART Hands framework is composed of two main stages: inference and alerting.

**Inference**
The inference stage is organized in four steps: multi-view capture, pose extraction, hand cropping, and CNN inference. The inference stage is illustrated in Figure 2.

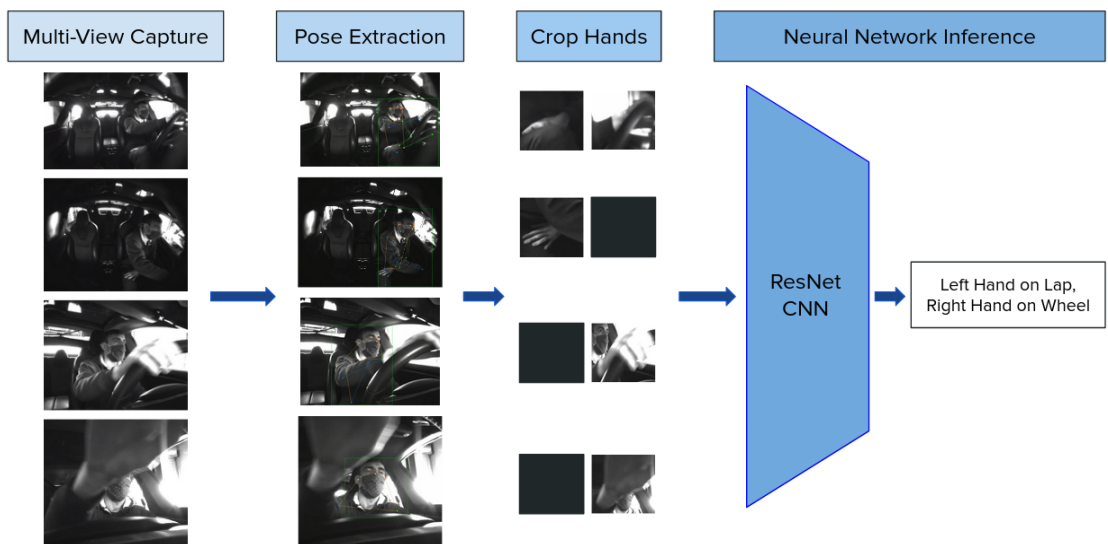

*Figure 2. The inference stage of the SMART Hands framework is conceptualized in four steps: multi-view capture, pose extraction, hand cropping, and CNN-based inference.*

**Multi-View Capture** The SMART Hands algorithmic framework performance depends on how completely the ensemble of cameras capture the driver's hands and surroundings. Various placements of the cameras are acceptable, and the framework is most effective when the views provide complementary information. For example, the views shown in Figure 3 capture overhead, straight, and side angles of the driver's hands. A multi-camera view can thoroughly capture hand activity and minimize obstructions that may be present with a single camera. In our example, further discussed in the Experiments section, a rearview mirror camera captures a side-facing angle on the hands that may be useful if the driver is holding an object in front of them, blocking the view of a dashboard camera. Two dashboard cameras are useful in capturing a front-facing view of the hands that can help if the driver has leaned or moved out of view of the rearview mirror camera. A camera behind the steering wheel is useful in capturing a close-up view of the driver's hands when their hands are on the wheel, and although it has a limited field of view, its limitations are compensated by the other cameras.

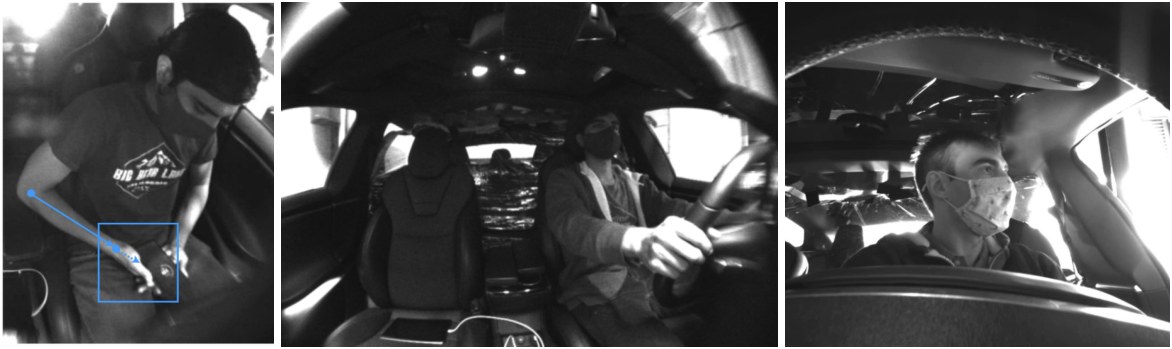

*Figure 3. Complementary views of multiple cameras can capture a variety of information about the driver's state.*

**Pose Extraction** Following data capture, we extract the pose of the driver in each frame, where "pose" is a collection of 2D keypoint coordinates associated with the driver's body, such as the wrists, elbows, shoulders, eyes, etc. This problem is broken into two steps: first, we must detect the driver in the frame, then detect the driver's pose. Each step requires its own neural network; for driver detection, we first use the Faster-RCNN [16] model with Feature Pyramid Networks [17], using a ResNet-50 backbone [18] to detect the driver. This network can operate at 28.8 frames per second, and we note that this network will output any humans detected in the frame, so we apply a post processing step (based on the camera view) to only include detections corresponding to the driver's seat. For joint detection, we employ the HRNet [19] model, which can operate at 22.7 frames per second.

These networks are evaluated using the mean Average Precision 50 metric. When considering all predictions made by the network, this tells us how many of those predictions have at least 50 percent overlap with the true human (or pose features). This can be thought of as similar to a "hit rate". Faster R-CNN, a widely-adopted model, can detect objects with .636 mAP, and HR-Net, a state-of-the-art pose estimator, can determine pose with .905 mAP. This portion of the inference system is the timing bottleneck, altogether able to run at approximately 15 frames per second, or 6-hundredths of a second per frame. The output of each stage of pose extraction are shown in Figure 4.

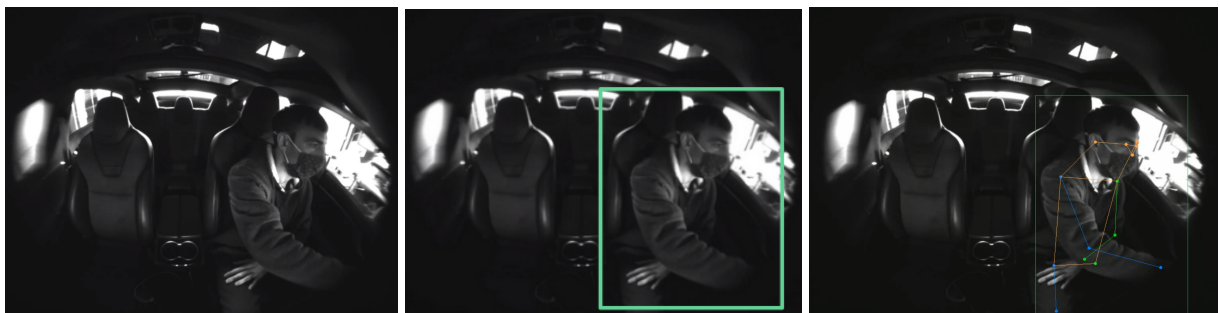

*Figure 4. The stages of pose extraction occur in two steps. First, the frame captured by a particular view (left) is fed to a human detection module (center). Then, joint keypoints associated with the detected human are determined (right).*

**Hand Cropping** We crop images around each of the hands, centered at the wrist and extending 100 pixels in each direction. This value is a hyperparameter which can be tuned for each camera view.

**CNN Inference** The cropped images are classified by a series of convolutional neural networks. The network is composed of four parallel sets of convolution layers (ResNet-50 backbones), which act on each of the four image views. Following the convolutional layers, each parallel track is fed to its own fully-connected layer of 512 nodes (followed by a ReLU activation). These layers are joined together by a fully-connected layer with 2048 nodes (followed by a softmax activation); this is the point of fusion, where the features extracted from the four views are combined and the relationships between the multiple views are learned.

This network outputs probabilities that the hands are holding one of three objects: Phone, Beverage, Tablet; or holding nothing. In the case that the hand is holding nothing, a second network (identical to the first) predicts the probability that the hand is in one of five hand location classes: Steering Wheel, Lap, Air, Radio, or Cupholder. Finally, the network infers the hands to be classified according to the class of maximal probability, completing the inference pipeline.

**Alerting**

Following inference, we apply additional steps to ensure that any alerts issued are meaningful, aiming to reduce noise and nuisance. The alerting stage is organized in three subsequent steps: post-processing, thresholding, and alerting, shown in Figure 5.

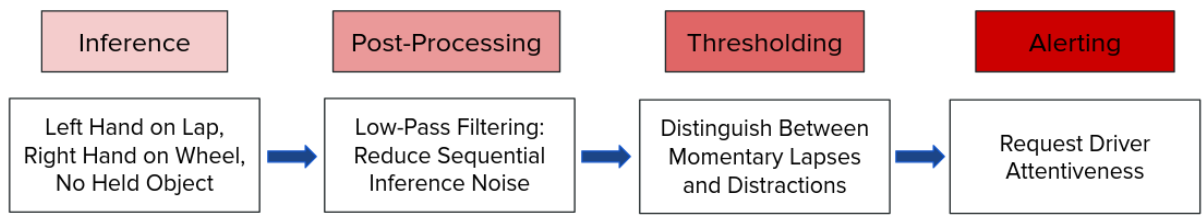

*Figure 5. The alerting stage of the SMART Hands framework. After inference, three steps are applied: post-processing, thresholding, and alerting.*

**Post-Processing** We employ low-pass filtering to reduce the effects of noisy patterns from inference (that is, small "blips" between classes for fractions of a second). This allows for a more steady prediction result by averaging over moving windows of time. We use a window of size 3, but this hyperparameter can be tuned based on observation of the duration of a typical inference mistake made by the network.

**Thresholding** We employ a thresholding step to distinguish between monetary lapses of attention (such as a driver quickly reaching for an object), versus an elongated period of distraction, which warrants an alert. We set our threshold to 150 (approximately 5 seconds at 30 frames per second). This hyperparameter should be tuned according to the goals of the automaker or driver policy.

**Alerting** If it is decided that the driver may be distracted, the system can then issue a standard request for driver attentiveness, or employ other downstream safety mechanisms. It is recommended that the alert system employs a method of alerting aligned to the standards of human-machine interface research; these techniques are outside the scope of this research, but we emphasize that this is a modular endpiece, and this framework can be applied for any downstream alerting mechanism.

**METRICS AND EVALUATION**

To evaluate SMART Hands, we created a 4-camera ensemble, illustrated in Figure 6. We use 4 Intel RealSense 435 Cameras enclosed in custom 3D-printed mounts, where two are placed on the central dashboard of the vehicle, one is located behind the steering wheel, and one is above the rearview mirror. These cameras contain an IR pass filter in order to account for the variability of real-world lighting conditions inside the vehicle and ensure robust detection of the hands in various conditions. We utilize this IR image in our experiments, but the framework could also extend to the RGB images or stereo image produced by the camera. The frame rate is approximately 30 frames per second, suitable for real-time data stream applications.

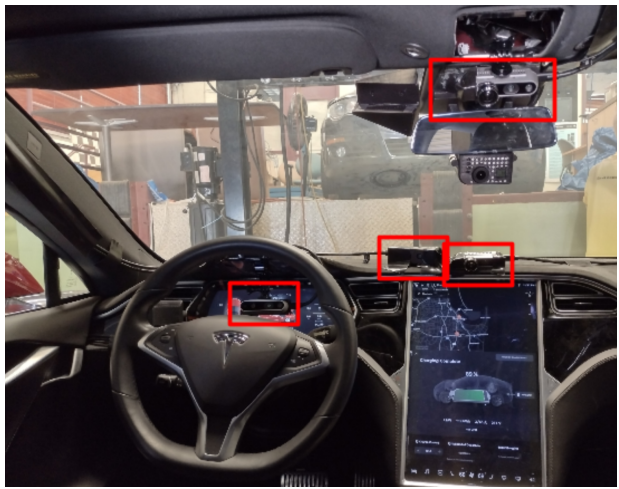

*Figure 6. The multi-camera configuration relative to the rest of the vehicle interior, featuring cameras located behind the wheel, on the dashboard facing the driver, on the dashboard facing center, and above the rearview mirror.*

Using this ensemble of cameras, we collect a dataset of 19 subjects engaged in various hand placements and object-related activities. Altogether, we collect approximately 81,000 frames corresponding to hand zone activity, and 128,000 frames corresponding to held object activity. We divide these into training, validation, and test sets using approximately 80%, 10%, and 10% of the data respectively (with marginal differences to account for dropped frames).

Using this data, we trained the SMART Hands neural networks for hand location and held object classification into the defined zones (3 location zones for the left hand, 5 location zones for the right hand, and 4 held objects [including null] for each hand).

**RESULTS**

Confusion matrices shown in Figures 7 and 8 describe the performance of the neural networks on classifying the hand location and held object to the defined classes. In all confusion matrices, the rows represent the true class, while the columns represent the predicted class. We show an average accuracy of 99.3% for left hand location, 99.2% for right hand location, 98.6% for left hand held object, and 99.2% for right hand held object. A qualitative example of SMART Hands' output is shown in Figure 9.

|  | Wheel | Lap | Air |
|---|---|---|---|
| Wheel | 4,699 | 2 | 13 |
| Lap | 13 | 2,322 | 7 |
| Air | 23 | 8 | 2,106 |

|  | Wheel | Lap | Air | Radio | Cupholder |
|---|---|---|---|---|---|
| Wheel | 3,558 | 0 | 6 | 1 | 0 |
| Lap | 18 | 1,920 | 2 | 0 | 0 |
| Air | 23 | 0 | 1,686 | 0 | 2 |
| Radio | 2 | 0 | 6 | 302 | 0 |
| Cupholder | 3 | 0 | 7 | 0 | 1,669 |

*Figure 7. Confusion matrix for left and right hand location (on left and right, respectively).*

|  | None | Beverage | Phone | Tablet |
|---|---|---|---|---|
| None | 9,215 | 27 | 5 | 0 |
| Beverage | 44 | 1,586 | 4 | 0 |
| Phone | 81 | 8 | 2,573 | 1 |
| Tablet | 19 | 10 | 11 | 1,902 |

|  | None | Beverage | Phone | Tablet |
|---|---|---|---|---|
| None | 9,318 | 20 | 1 | 2 |
| Beverage | 29 | 1,596 | 2 | 15 |
| Phone | 14 | 4 | 1,977 | 0 |
| Tablet | 15 | 7 | 2 | 1,622 |

*Figure 8. Confusion matrix for left and right hand held objects (on left and right, respectively).*

**CONCLUSIONS**

Hand activity monitoring via a multi-camera configuration has a strong potential impact on traffic safety. Approximately, 830,000 Tesla Autopilots, 1.5 million Subarus, 1.3 million Hyundais, and 200,000 Cadillacs with either driver-facing IR cameras or hand-monitoring steering wheel sensors are on the road [20]. These 4.3 million vehicles with driver-facing assistance systems represent approximately 1.5% of the 287 million vehicles in the fleet. Under the same rate of commercial growth exhibited in the past five years [21], we expect, as a lower bound, that 3% of all fleet vehicles will contain these systems in the next 5 years. Because Subaru's EyeSight system for eye monitoring has reduced accidents by 85% [22], and recent research finds the hands to be even more safety-critical than the eyes in assessing driver takeover readiness, we suggest a robust, multi-camera configuration for hand activity monitoring will improve reduce crashes by over 90% in vehicles that currently have driver-facing assistance systems, therefore by 2.7% in the fleet. Distracted driving led to about 680,000 accidents in 2020 [23], so we estimate this configuration may prevent 18,360 (2.7%) of those accidents.

**Applications and Demonstration**
The multi-camera framework presented here has numerous ancillary and downstream applications. Figure 9 illustrates the immediate demonstrable output of SMART Hands as capable of inferring information on hand activity from the camera streams. As the Faster-RCNN module detects all individuals in the car, it can also be used to alert the driver of distracting behaviors by the passengers, if a child has been left behind in the car [24] or even if safety rules are not being followed by the occupants, such as failure to wear seatbelts or poor seat positioning [25], critical in pre-crash airbag deployment applications. The pose of the driver captures all the keypoints - the head, the eyes and the hand - which allows this work to be used in conjunction with driver drowsiness and distraction using gaze and head poses, or computation of metrics like Observable Readiness Index and Take-Over Time [4, 6].

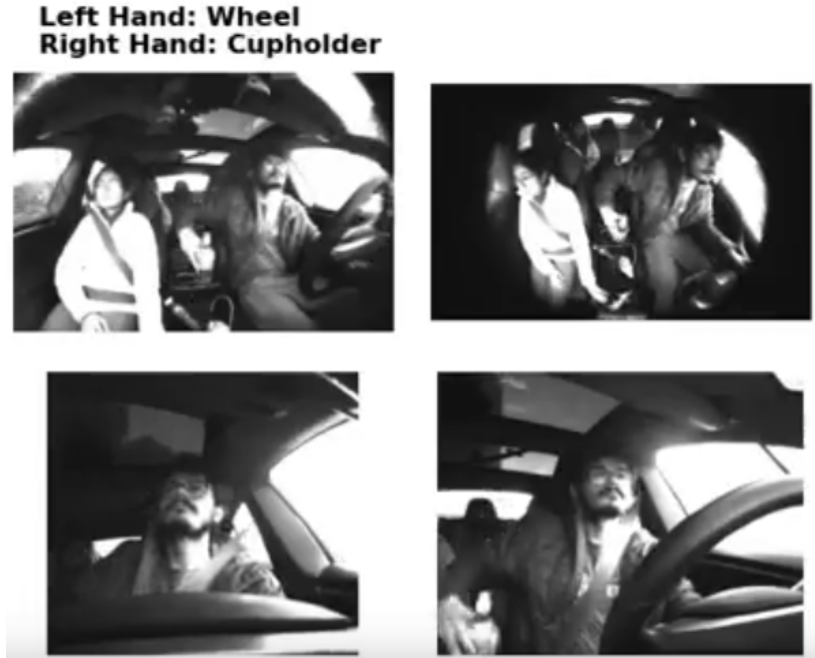

*Figure 9. SMART Hands independently infers the hand activity for the driver's left and right hand.*

As downstream applications, we envision this framework to easily scale as an aftermarket product with models trained to generalize on a wide variety of cross-validated subjects, cars, and atypical scenarios (e.g. drivers with gloves, food, etc.). Self-supervised and active learning create further potential for system predictions to continuously improve models [24]. Stereo vision can also exploit the known configuration of cabin structures in different cars to generalize better (in fact, stereo vision is already present in the aforementioned RealSense cameras). A few examples of the end product could be (1) issuing alerts when sustained distraction is identified to refocus the driver's attention, (2) preventing engagement of autopilot or cruise control if the driver is distracted, and (3) autonomous cars deciding whether to cede control in critical situations when the driver may not be able to react due to their distraction.

SMART Hands can be experienced in live, real-time demonstration. In this setup, participants would be seated in a simulator and have multiple cameras recording their hand activities (eating, drinking, using their phones, etc.). SMART Hands would display the detected hand activity or hand zone, and if non-driving activities are sustained, issue alerts for the driver to put their hands back on the steering wheel, with a visual example in Figure 10.

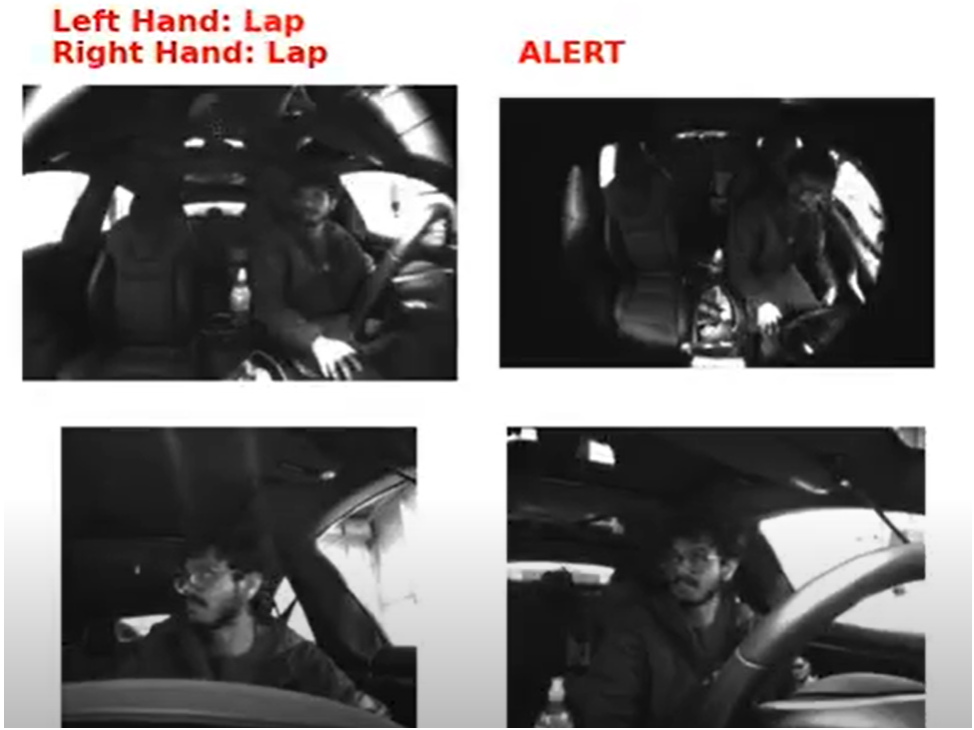

*Figure 10. The SMART Hands framework analyzes sequences of images to identify periods of sustained non-driving activity, allowing systems to issue advisory alerts to the driver.*